\documentclass[sigconf]{acmart}
\AtBeginDocument{%
  \providecommand\BibTeX{{%
    \normalfont B\kern-0.5em{\scshape i\kern-0.25em b}\kern-0.8em\TeX}}}

\copyrightyear{2021}
\acmYear{2021}
\setcopyright{acmcopyright}\acmConference[CIKM '21]{Proceedings of the 30th ACM International Conference on Information and Knowledge Management}{November 1--5, 2021}{Virtual Event, QLD, Australia}
\acmBooktitle{Proceedings of the 30th ACM International Conference on Information and Knowledge Management (CIKM '21), November 1--5, 2021, Virtual Event, QLD, Australia}
\acmPrice{15.00}
\acmDOI{10.1145/3459637.3482021}
\acmISBN{978-1-4503-8446-9/21/11}

\usepackage{flushend}
\usepackage{balance}

\usepackage{todonotes}
\presetkeys{todonotes}{inline}{}

\usepackage{tabularx}
\usepackage{enumitem}
\usepackage{dirtytalk}
\usepackage{hyperref}
\hypersetup{
    colorlinks=true,
    linkcolor=blue,
    filecolor=magenta,      
    urlcolor=cyan,
}

\usepackage{graphicx}
\usepackage{xcolor}
\usepackage{soul}

\usepackage[ruled,vlined,noend]{algorithm2e}

\begin{document}

\title{VerbCL: A Dataset~of~Verbatim~Quotes for Highlight~Extraction~in~Case~Law}

\author{Julien Rossi}
\email{j.rossi@uva.nl}
\orcid{0000-0003-0548-6102}
\affiliation{%
  \institution{University of Amsterdam}
  \country{Netherlands}
}

\author{Svitlana Vakulenko}
\email{s.vakulenko@uva.nl}
\affiliation{%
  \institution{University of Amsterdam}
  \country{Netherlands}
}

\author{Evangelos Kanoulas}
\email{e.kanoulas@uva.nl}
\affiliation{%
  \institution{University of Amsterdam}
  \country{Netherlands}
}

\renewcommand{\shortauthors}{Rossi et al.}

\begin{abstract}
Citing legal opinions is a key part of legal argumentation, an expert task that requires retrieval, extraction and summarization of information from court decisions. The identification of legally salient parts in an opinion for the purpose of citation may be seen as a domain-specific formulation of a highlight extraction or passage retrieval task. As similar tasks in other domains such as web search show significant attention and improvement, progress in the legal domain is hindered by the lack of resources for training and evaluation. 
This paper presents a new dataset that consists of the citation graph of court opinions, which cite previously published court opinions in support of their arguments. In particular, we focus on the verbatim quotes, i.e., where the text of the original opinion is directly reused.
With this approach, we explain the relative importance of different text spans of a court opinion by showcasing their usage in citations, and measuring their contribution to the relations between opinions in the citation graph.
We release VerbCL\footnote{\url{https://github.com/j-rossi-nl/verbcl}}, a large-scale dataset derived from CourtListener and introduce the task of highlight extraction as a single-document summarization task based on the citation graph establishing the first baseline results for this task on the VerbCL dataset.
\end{abstract}



\maketitle

\begin{table*}[ht]
    \begin{tabular}{|p{0.11\textwidth}|p{0.85\textwidth}|}
        \hline
        Citing opinion & \textbf{ID}: 2346822\\
        & \textbf{Title}: LL Ex Rel. Doe v. Chimes Dist. of Columbia, Inc\\
        & \textbf{Text}: 
        
        (...) a special relationship that gave rise to a duty on the part of the United States to protect or warn her.[1] There was no such special relationship between L.L. and the United States, however, and the United States motion must accordingly be granted. \hl{"Ordinarily, the owner or possessor of land is under no duty to protect invitees from assaults by third parties while the invitee is upon the premises ... [unless] there is a special relationship between [the] possessor of land and his invitee giving rise to a duty to protect the invitee from such assaults."} Wright v. Webb, \colorbox{pink}{CITATION\_1239944}, 920-21 (1987). Chimes submits that this is a "special relationship" case. The cases on which Chimes relies for that proposition presented distinct factual bases for finding a "special relationship."  (...) \\

        \hline
        Cited opinion & \textbf{ID}: \colorbox{pink}{1239944}\\
        (score $=1.71$) & \textbf{Title}: Wright v. Webb \\

        & \textbf{Text}: 
        
        (...) We will assume, without deciding, that Webb was the Wrights' business invitee. Thus, the Wrights owed Webb the duty of ordinary care to maintain their parking lot in a reasonably safe condition. See Tate Rice, 227 Va. 341, 345, 315 S.E.2d 385, 388 (1984).

        \hl{Ordinarily, the owner or possessor of land is under no duty to protect invitees from assaults by third parties while the invitee is upon the premises.} Restatement (Second) of Torts | 314A (1965) recognizes exceptions to the rule of non-liability for the assaults of a third party where \hl{there is a special relationship between a possessor of land and his invitee giving rise to a duty to protect the invitee from such assaults.} We alluded to this Restatement rule in both Klingbeil Management Group Co. Vito, 233 Va. 445, 447, 357 S.E.2d 200, 201 (1987) and Gulf Reston, Inc. Rogers, 215 Va. 155, 158, 207 S.E.2d 841, 844 (1974), but made it plain in Gulf Reston that this was only a reference to the Restatement rule. Our statement in Klingbeil was simply a comment upon the reference in Gulf Reston. 
        (...) \\
        \hline
    \end{tabular}
    \caption{A sample of a verbatim quote from the VerbCL dataset.}
    \label{table:sample}
\end{table*}

\section{Introduction}

Courts of common law systems rely on statutes and precedents for deciding the law applicable to a case. A court opinion on a case, being a decision taken by a court when litigating a specific case,  is a text assembled by the court judges using the parties' arguments and rebuttals of arguments. Previously published opinions (also known as case law) are cited to support the argumentation of the parties and the opinion of the court on the litigated case.

Thereby a legal discourse is made of a mix of factual details from the case as well as legal discussions arising from the qualification of the facts, driven by the past discussions with a similarity, either in facts or in reasoning. We provide an example, taken from the VerbCL dataset that illustrates how case law citation is used in practice, in Table~\ref{table:sample}.

For any legal-domain practitioner, search for an appropriate citation is a key task in preparing the argumentation for a case, which needs to be facilitated by an information retrieval system. This task was considered in case law retrieval~\cite{coliee2018, coliee2019} as an information retrieval task where the query is a draft of an opinion and the result is a ranked list of past opinions. Relevance in this case may be guided by the similarity between the query and the retrieved opinion, or by the fact that the retrieved opinions address legal facets or questions in the query, as well as by the consideration whether the retrieved opinions might be worth being cited or not in the context of the query. 

In this work, we introduce the new task of \textit{highlight extraction}: given the text of an opinion, predict the subset of text spans that are likely to be cited in the future. 
In this task, the question is whether we can successfully predict which parts of an opinion will have an impact on future litigation, i.e., will be cited.

The motivation behind this task is to facilitate the work done by a legal commentator, who is tasked to identify interesting points in an opinion at the time of its publication. The legal commentator inspects all recently published opinions looking for potentially valuable excerpts that are worth being cited. The evaluation whether a part of an opinion is of a particular interest, or might be of interest in future cases is informed by the knowledge of the jurisprudential landscape, so the commentator can contrast in a new opinion what makes part of this landscape, and what is novel.

We aim to inform the task of highlight extraction through the construction of a large-scale dataset that will allow the development and evaluation of data-driven models able to solve the task. To this end, we use a public dataset of US court opinions, CourtListener\footnote{\url{https://www.courtlistener.com}}, to extract a citation network of court opinions.

The citation network naturally emerges from the repository of opinion documents since each opinion usually quotes multiple previously published opinions for the purpose of the legal argumentation of the case.
We consider this network as a directed acyclic graph, where each node stands for an opinion document.
Edges represent the citations made in opinions, directed from the citing opinion's node towards the cited opinion's node. 
In this paper, we consider the citing-cited relation as follows: a citing opinion is making an argument by referring to a cited opinion.

In the text of the citing opinion, a citation is introduced by an \textit{anchor}, that we consider to be the supported legal argument.
More specifically, we differentiate between two types of anchors: abstractive anchors, where the argument is cited as a paraphrase of the original opinion document, and extractive anchors (or \textit{verbatim quotes}) where the argument is directly extracted from the cited opinion's text by copying the relevant text span.
We focus on the verbatim quotes in this work since they are much easier to detect and reproduce than paraphrases.

We further consider that verbatim quotes emphasize which part of the cited opinion have a legal weight, bringing novelty and importance into the legal landscape at the time of the publication of an opinion document.
We define a \textit{highlight} of an opinion as the set of all text spans that are later cited as verbatim quotes.
Therefore, we can produce the highlights for the previously published opinions by considering the citation graph.
The highlight of an opinion can be constructed by considering all the verbatim quotes of this opinion used in the citing opinion documents (see an example of the highlight in Table~\ref{table:sample}).

\begin{figure}[h]
    \centering
    \includegraphics[width=\linewidth]{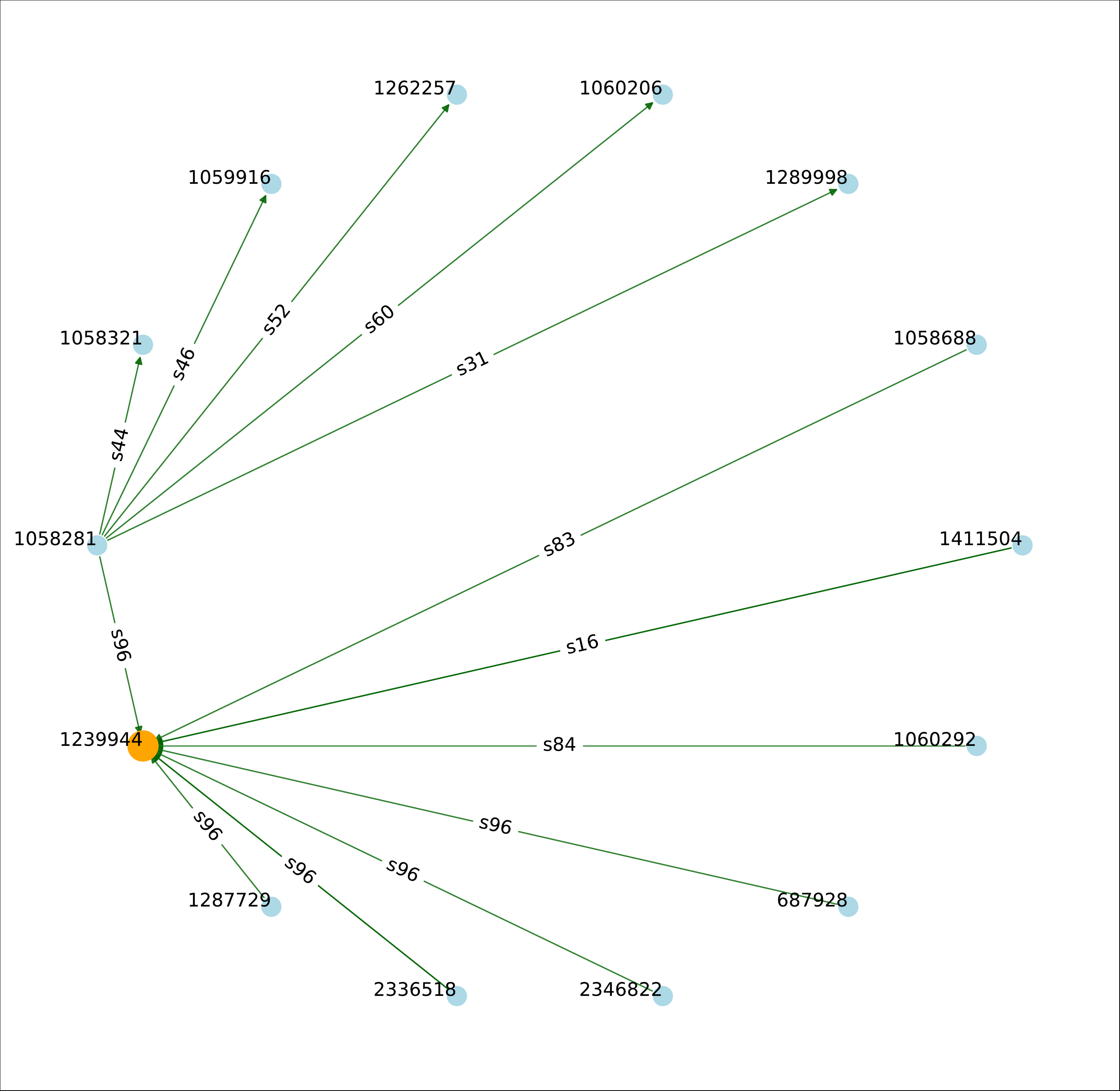}  
    \caption{Citation graph, where each node is an opinion document with edges pointing from a citing opinion to a cited opinion. Every edge is labeled with the identifier of a sentence of the cited opinion that was quoted verbatim. For example, sentence 96 of opinion 1239944 is cited by the opinion 1058281, which contains sentence 44 cited by opinion 1058321.}
    \label{fig:citation-graph}   
    
\end{figure}

To aggregate information about all citing opinions, the citation graph is used.
We illustrate the citation graph around the cited opinion from Table~\ref{table:sample} in Figure~\ref{fig:citation-graph}, which shows all the citing opinions as well as the identifiers of their verbatim quotes as the edge labels.

The goal of highlight extraction is to predict the highlights for new opinions, i.e., predicting the future citations based on the opinion text.
We proceed by establishing the first baseline for the highlight extraction task using the state-of-the-art single-document summarization approaches: TextRank~\cite{barrios2016variations} and PreSumm~\cite{presumm}, as well as a sentence-level classifier based on DistilBERT~\cite{sanh2020distilbert} . The goal of our experiments is to verify to which extent the highlight extraction task in the citation network can be addressed using single-document summarization approaches without any information about the citation graph structure.

\begin{samepage}
    Our main contributions are two-fold:
    \begin{enumerate}
        \item the dataset with a citation graph, verbatim quotes and opinion highlights;
        \item the task of highlight extraction with baselines.
    \end{enumerate}
\end{samepage}

We start by describing how VerbCL was constructed in Section~\ref{section:dataset-construction} and discuss its main characteristics in Section~\ref{section:dataset-statistics}. 
We formalize the task of highlight extraction and report on the experiments we conducted with the baseline models for this task in Section~\ref{section:highlight-extraction}. The overview of existing datasets for case law retrieval, and related work on highlight extraction and citation-based summarization in other domains is given in Section~\ref{section:related-work}. We conclude and propose directions for future work in Section~\ref{section:conclusion}.

\begin{table*}[h]
    \begin{tabularx}{\linewidth}{|l|l|X|}
        \hline
        Field Name & Type & Comment \\
        \hline
        \texttt{citing\_opinion\_id} & \texttt{int} & The \texttt{opinion\_id} of the \textbf{CITING} opinion. \\
        \texttt{cited\_opinion\_id} & \texttt{int} & The \texttt{opinion\_id} of the \textbf{CITED} opinion. \\
        \texttt{sentence\_id} & \texttt{int} & The \texttt{sentence\_id} of the source sentence in the CITED opinion. \\
        \texttt{verbatim} & \texttt{str} & The span of text in the CITING opinion that we consider to be a potential verbatim quote from the CITED opinion \\
        \texttt{snippet} & \texttt{str} & The span of text from the CITING opinion around the in-text citation of the CITED opinion. It contains 100 words before the in-text citation, and 100 words after the in-text citation. \\
        \texttt{score} & \texttt{float} & The score of \texttt{verbatim} for being an actual verbatim quote from the CITED opinion, placed in the CITING opinion. A score of $-1$ indicates our model does not consider it as being a quote from the CITED opinion. \\
        \hline
    \end{tabularx}
    \caption{Fields of the documents in VerbCL Citation Graph.}
    \label{table:citation-graph-fields}
\end{table*}

\begin{table*}[h]
    \begin{tabularx}{\linewidth}{|l|l|X|}
        \hline
        Field Name & Type & Comment \\
        \hline
        \texttt{opinion\_id} & \texttt{int} & The unique identifier of the opinion. \\
        \texttt{sentence\_id} & \texttt{int} & The unique identifier of the sentence: its index within the list of sentences of the opinion. \\
        \texttt{raw\_text} & \texttt{str} & The complete text of the sentence. \\
        \texttt{highlight} & \texttt{bool} & The binary label of the sentence. $True$ means the sentence has been quoted verbatim in a citing opinion. \\
        \texttt{count\_citations} & \texttt{int} & The number of times the sentence has been quoted verbatim. \\
        \hline
    \end{tabularx}
    \caption{Fields of the documents in VerbCL Highlights.}
    \label{table:highlights-fields}
\end{table*}

\section{Dataset Construction}
\label{section:dataset-construction}

VerbCL is based on the Court Listener dataset, a collection, which is publicly available for free.
In this section, we describe the process of constructing the dataset. 

\subsection{Court Listener}
CourtListener is a project of the Free Law Project\footnote{\url{https://free.law/}}, a US non-profit organization started in 2010, with the aim of \say{making the legal world more fair and efficient}.

The original Court Listener dataset is a collection of every court opinion published by every court in the United States. It covers 406 jurisdictions (out of 423), with opinions from the year 1754 up to now. It is constantly updated with newly filed opinions, and digitized archives. 

We obtained the Court Listener dataset by downloading the bulk data on September 1st 2019.\footnote{\url{https://www.courtlistener.com/api/bulk-info/}}. More recent versions will have more opinions available, the whole processing will apply as long as the underlying relational database scheme remains unchanged. 

The unit of data in Court Listener is the court opinion. Each court opinion is a single unique decision made by a specific court at a specific date on a specific case. A single case can be litigated by many courts, in case of appeals for example, so opinions are clustered per case. The Court Listener dataset follows the scheme of a relational database, where each opinion belongs to one cluster. Opinions are also clustered in dockets, a single docket gathering the different opinions made by a unique court on a unique case, as it happens that the same court may emit multiple opinions on the case at different dates.

At a more granular level, Court Listener represents each opinion as an HTML document, where specific tags mark where an in-text citation of case law appears in the opinion. Free Law implements a citation parsing and matching program\footnote{https://github.com/freelawproject/courtlistener} that allows for an automated annotation of the unstructured text of the original court opinion. 

\subsection{Data Processing}

For the construction of VerbCL, we focus on the opinions of CourtListener whose full text is available as HTML code including tagged in-text citations (looking for the field \texttt{html\_with\_citations}). The HTML code for each opinion has specific tags to identify citations made in the text of the opinion, each citation being identified with the unique identifier of the cited opinion. 

We consider that each citation is introduced by an anchor, which is located in the vicinity of the citation tag. The anchor is the ``raison d'être'' of the citation, it is an argument to the current case, it is participating to the litigation of the case, and its validity is affirmed by invoking a similar argument that has already been made in an existing court opinion. 

We focus specifically on the anchors that are extracts of the text of the cited opinion (\textit{verbatim quotes}), similar to the example provided in Table~\ref{table:sample}. 

\begin{algorithm}[htbp]
\SetAlgoLined
\KwResult{Verbatim Quotes}
 initialization\;
 \ForAll{opinions}{
    \ForAll{citations}{
        snippet = N words before / after in-text citation\;
        candidates = potential verbatim quotes\;
        \ForAll{candidates}{
            \If{candidate is verbatim}{
                identify original sentences in cited opinion\;            
                store data point\;
            }
        }
    }
 }
 \caption{Data Processing Pipeline}
\end{algorithm}

Our data processing procedure includes extracting the text around the in-text citation, identifying potential verbatim quotes (often marked by quotation marks) and eventually classifying those who actually are verbatim quotes from the text of the cited opinion. Our code includes a framework for parallel distributed processing of massive dataset, using \texttt{pyarrow}, Elasticsearch\footnote{\url{https://www.elastic.co/}} through its Python API, and MongoDB\footnote{\url{www.mongodb.com}} together with \texttt{pymongo}.

\subsubsection{Snippets extraction}

Although the anchor is in the vicinity of the in-text citation, we observe a wide range of variations in the presentation and wording of these anchors, in relation to the in-text citation. Our strategy is to reduce this problem  to identify a verbatim quote in the direct surroundings of the in-text citation, defined as the N words before and after the citation. We opted for a value $N=100$, which is a compromise between high recall (the more text around the citation we consider, the more likely it is we will identify the anchor), and computational footprint. Experiments included: considering entire paragraphs, but it yielded too much text; identifying nearby sentences, but the high numbers of acronyms and abbreviations around in-text citations renders sentence splitting useless.

\subsubsection{Verbatim candidates}

The identification of potential verbatim quotes is entirely rule-based, around the usage of different quote characters. Each snippet will generate multiple spans that were enclosed between those quote characters.

\subsubsection{Qualifying verbatim quotes}

At this stage, we want to confirm for each verbatim candidate whether it is a text that originates from the cited opinion. In the context of studying a specific in-text citation in a citing opinion, the cited opinion is known, so we have to evaluate whether a text span is lifted from the full text of the cited opinion.

The problem of identifying segments of text within a long text has proven to be difficult. As most of the opinions originate from printed books, OCR artifacts are expected, such as misplaced spaces or misspellings. This will affect the quality of the text in both cited and citing opinion. The common practice of using ellipsis in the citing opinion's text (as can be seen in the sample shown in Table~\ref{table:sample}) will result in the suppression of fragments of the original text. 

Traditional approaches to this fuzzy matching problem rely on heuristics based on the computation of editing distances, such as the Levenshtein Distance~\cite{1966SPhD...10..707L}. The design of the heuristics is a first challenge, given the sheer variety of differences that could be observed between the original text in the cited opinion and its counterpart in the citing opinion. More importantly, the computational footprint of calculating an editing distance renders this method inapplicable to our dataset.

We built a classifier based on ``Interval Queries'' offered by Elasticsearch. We refer the reader to the provided source code for the exact parameters of this query. The classifier predicts that a span of text belongs to the positive class when the query returns a non-empty result, and predicts the negative class when this query does not produce any search result. We manually annotated a test dataset for this classifier which revealed it had high precision and recall for both classes. The reader can refer to Chapter~\ref{subsection:dataset-quality}.

As a result of this stage, we have built the VerbCL citation graph, a directed graph where nodes are opinions and edges are verbatim quotes.

\subsubsection{Identifying highlights}

Highlights in the cited opinion are the sentences that were used for verbatim quotes. For each span of text identified as a verbatim quote of the cited opinion, we identify which sentence or sentences, are actually quoted from the text of the cited opinion. This is achieved through another set of Elasticsearch queries that are run over an indexed collection of opinion sentences. For sentence tokenization, we used the Punkt sentence tokenizer from \texttt{NLTK}~\cite{bird2009natural}.

As a result, we have built the VerbCL Highlights dataset, a dataset of opinions annotated at sentence level with a binary label where the positive class is made of sentences that were later on cited as verbatim quotes.

\subsubsection{Dataset structure}

Following this stage, we have produced the following data collections:
\begin{itemize}[]
    \item \textbf{VerbCL Highlights}: An annotated collection of all opinion sentences, with a binary label for which True indicates that the sentence was cited verbatim;
    \item \textbf{VerbCL Citation Graph}: An annotated citation graph for all case law citations with a verbatim quote of the cited opinion. It is a directed acyclic graph $(V, E)$ where $V$ is the set of all opinion documents and $E$ is the set of verbatim quotes.
\end{itemize}

VerbCL Highlights is used to inform the highlight extraction task since it contains the original opinion text and the subset we consider as the correct highlight.
VerbCL Citation Graph was used to produce VerbCL Highlights and provides auxiliary information about the opinion text reuse that can be used for the highlight extraction task.

See Tables~\ref{table:citation-graph-fields}-\ref{table:highlights-fields} for the structure of these subsets.
A tutorial on loading the data is available as a \texttt{Jupyter Notebook} in the code repository\footnote{\url{https://github.com/j-rossi-nl/verbcl}}.

\subsection{Quality Assurance}
\label{subsection:dataset-quality}

We manually annotated 180 random anchors, coming from random opinions, and evaluated whether or not the candidate snippet was an actual verbatim quote out of the cited opinion. The code for this manual annotation task is included in our codebase, it is based on the \texttt{Django} framework.\footnote{\url{https://www.djangoproject.com/}}

On this random sample of 180 texts, with 60 of them from the positive class (actual verbatim quotes from the cited opinion), our binary classifier had a Precision $P>0.96$ and Recall $R>0.92$ for both classes. We consider our solution for this problem to offer a sufficiently good compromise between computation and accuracy.

\section{Dataset Statistics}
\label{section:dataset-statistics}

In the following we describe the main characteristics of the VerbCL Citation Graph and the VerbCL Highlights.

\subsection{VerbCL Citation Graph}

From Court Listener, we consider the subset of opinions that either cite other opinions with a verbatim quote, or are cited verbatim by other opinions. This subset contains circa 1.5M opinions.

The main characteristics of our dataset are summarized in Table~\ref{table:stats}. Considering that we are studying the nodes of a citation network, we also disclose our analysis of the corresponding graph. 

\begin{table}[h]
    \begin{tabularx}{\linewidth}{|X|r|}
        \multicolumn{2}{c}{\textbf{Court Listener (CL)}} \\
        \hline
        Opinions & 4,265,231 \\
        \hline 
        Citing opinions & 3,062,334 \\
        Cited opinions & 2,020,779 \\
        Citations & 30,318,321 \\
        \hline
        \multicolumn{2}{c}{} \\
        \multicolumn{2}{c}{\textbf{VerbCL Citation Graph}} \\
        \hline
        Opinions & 1,493,561 \\
        & 35\% of opinions in CL \\
        \hline 
        Verbatim quotes & 6,210,703  \\
        & 20\% of citations in CL \\
        Verbatim citing opinions & 1,086,238 \\
        & 35\% of citing opinions in CL \\
        Verbatim cited opinions & 946,962 \\
        & 47\% of cited opinions in CL \\
        \hline
        Edges in the citation graph & 4,002,137 \\
        Graph density & $1.8 \times 10^{-6}$ \\
        \hline
        Number of words in a verbatim & $[5-100]$ \\
        Average & 15 \\
        Quartiles 25-50-75 & 12 - 20 - 30 \\
        \hline
    \end{tabularx}
    \caption{Dataset statistics of Court Listener (CL) and VerbCL.}
    \label{table:stats}
\end{table}

In the citation network, nodes are opinions and directed edges materialize the citations, from the citing opinion towards the cited opinion.
Thereby, the in-degree of a node $v$ is the number of times this opinion is cited by other opinions, while the out-degree indicates the number of opinions that the opinion corresponding to the node $v$ cites.

\begin{figure}
    \centering
    \includegraphics[width=\linewidth]{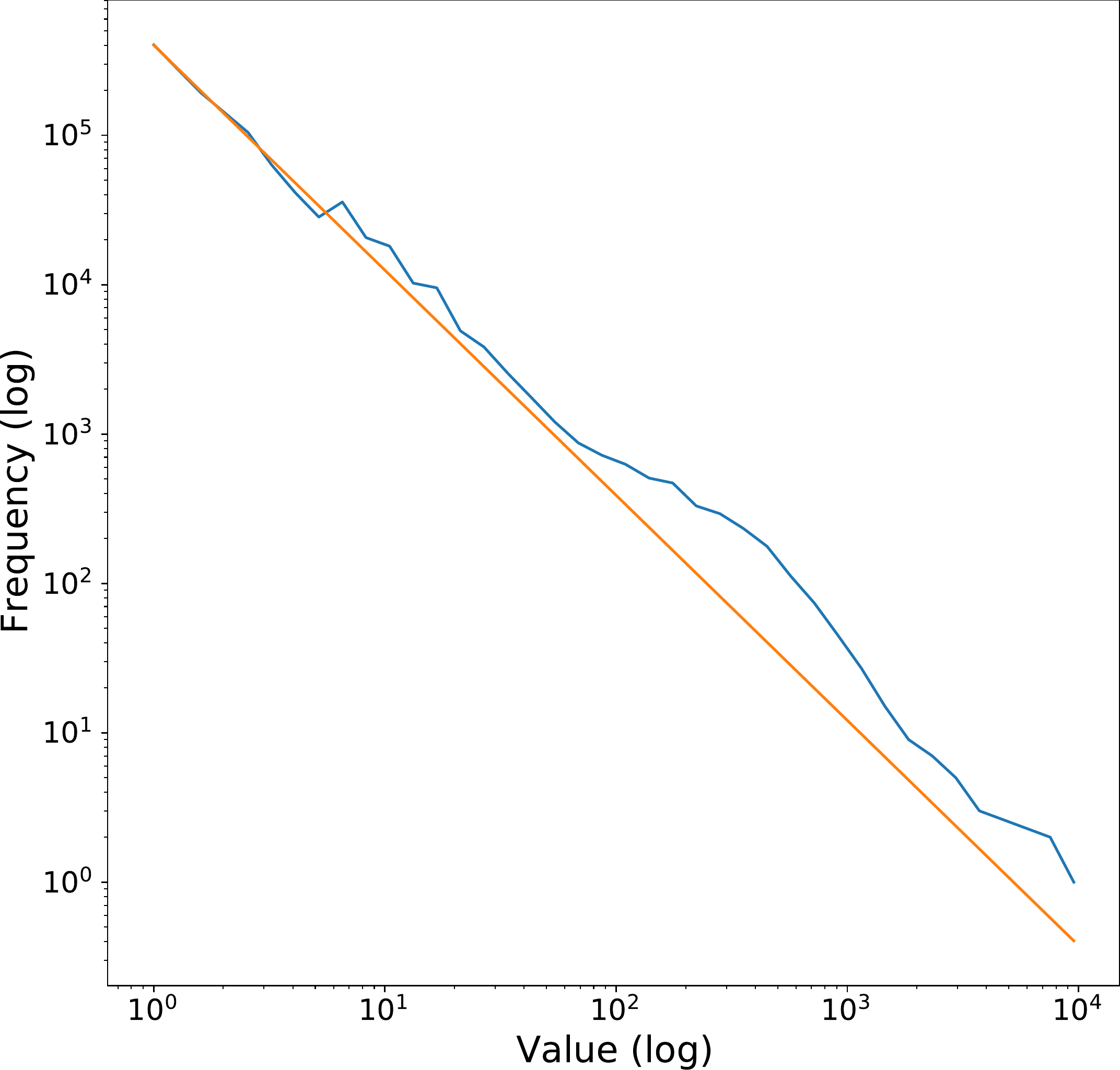}   
    \caption{Distribution of number of citations for an opinion.}
    \label{fig:distrib_indegree}   
\end{figure}

We plot the distribution of in-degree of the nodes in the citation network in Figure~\ref{fig:distrib_indegree}, as well as the Zipf's law, a power law with an exponent $k=-1$, which is a common distribution in the field of text mining. This shows us that these verbatim quotes have a behavior closer to the appearance of a term in a collection of documents, than to the expected distribution of links in a web graph (a power law with exponent $k \approx 2.1$).

We imported the citation network, using \texttt{networkx}~\cite{networkx}, in order to compute some descriptive centrality statistics of the graph. For computing time reasons, we used approximations instead of the actual values, following the experiments in~\cite{brandes2007centrality}.

\begin{figure}
    \centering
    \includegraphics[width=\linewidth]{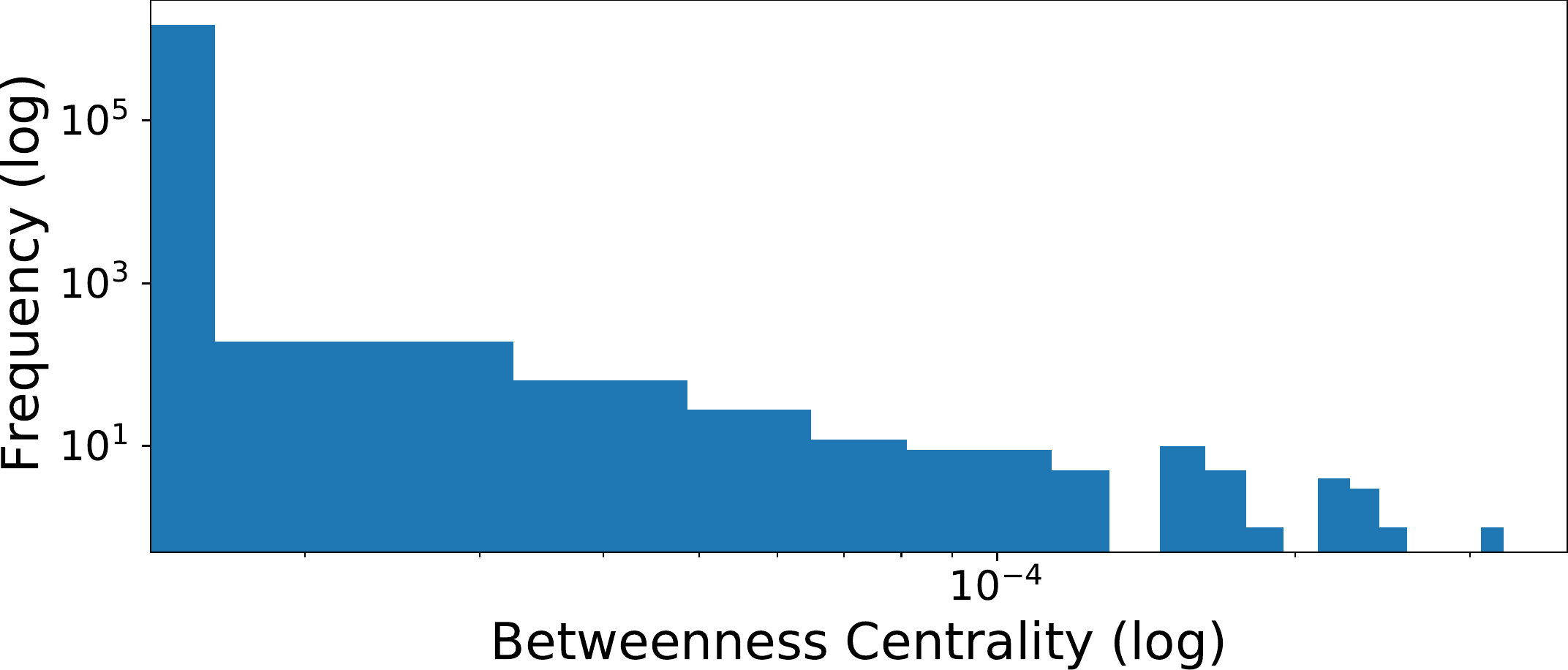}
    \caption{Distribution of betweenness centrality.}
    \label{fig:distrib_btwctr}
\end{figure}

The distribution of betweenness centrality of the nodes of the citation graph is plotted in Figure~\ref{fig:distrib_btwctr}, where we observe that the network contains a few very central nodes, corresponding to the group with the highest values.
A node with a high betweenness centrality is a node which is on the shortest path between many nodes of the graph, reflecting its importance in the network.
Betweenness centrality~\cite{brandes2001faster} of a node $v$ is defined as the sum of the fraction of all-pairs shortest paths that pass through $v$, see Equation~\ref{equation:betweenness-centrality}.
Since the exact calculation of betweenness centrality for every node of the network is too computationally expansive, we made use of an approximation algorithm~\cite{brandes2007centrality}.

\begin{equation}
    c_B(v) =\sum_{s,t \in V} \frac{\sigma(s, t|v)}{\sigma(s, t)}
    \label{equation:betweenness-centrality}
\end{equation}

When applied to academic literature and the corresponding citation network, betweenness centrality is considered to measure importance of a paper in the field, i.e., seminal academic papers were reported to have a relatively high betweenness centrality~\cite{leydesdorff2018betweenness,ORTEGA2014728}. 
In contrast, our citation network has only a few nodes with high relative betweenness centrality but the absolute values are still very low (maximum: $3 \times 10^{-4}$).

We hypothesize that this difference can be explained by much higher content redundancy in the court opinions in comparison with academic literature.
Novelty and originality of contributions are the main characteristics for an academic paper, which makes the paper take a unique and irreplaceable place in the network.
In contrast, courts of the justice system serve any valid case presented regardless of its originality.
Therefore, it is only natural that multiple cases can often make for equally good candidates in support of the same argumentation line.
This has an effect of the more even citation distribution across the whole network of court opinions in comparison with an academic network, which also reduces the centrality measures for court opinions as our data confirms.

Comparing the ranking of nodes by centrality with the ranking by outdegree (i.e. the number of times this opinion is cited by others), we compute the Kendall's Tau and Spearman's Rho coefficients of rank correlation: $\tau = 0.21, \, \rho = 0.27$, both with p-value $p = 0.0$, showing weak but significant correlation. We attribute this weak correlation to the historicity factor: contrary to  a webpages network, each node in our citation network is constrained by the capacity to cite only past opinions; we also consider the novelty factory, where a legal practitioner is biased towards most recent opinions when citing, therefore pushing the ``central'' opinions out of the shortest paths.

\subsection{VerbCL Highlights}

The main characteristics of our dataset are summarized in Table~\ref{table:stats-highlights}. These statistics come from a random sample of circa 10k opinions.

Sentences from one opinion that are cited verbatim by other opinions are considered as highlights of the cited opinion. The task of identifying the highlights from the full text of an opinion is introduced in this paper under the name ``Highlight Extraction'', in Section~\ref{section:highlight-extraction}.

\begin{table}[h]
    \begin{tabularx}{\linewidth}{|X|r|}
        \multicolumn{2}{c}{\textbf{VerbCL Highlights}} \\
        \hline
        Opinions & 1,493,561 \\
        \hline 
        Number of sentences per opinion & $99.5\%$ under $5000$ sentences \\
        Average & 437 \\
        \hline
        Number of tokens per sentence & $99.9\%$ under $171$ tokens \\
        Average & 24 \\
        \hline
        \% highlight sentences (per opinion, the number of highlights divided by the number of sentences) & $99.5\%$ under $21.8\%$ \\
        Average & $3.5\%$ \\
        \hline
        Number of citations per highlight & $99\%$ under $434$ citations \\
        Average & 34 \\
        \hline
    \end{tabularx}
    \caption{Dataset statistics for VerbCL Highlights.}
    \label{table:stats-highlights}
\end{table}

\begin{figure}[h]
    \centering
    \includegraphics[width=\linewidth]{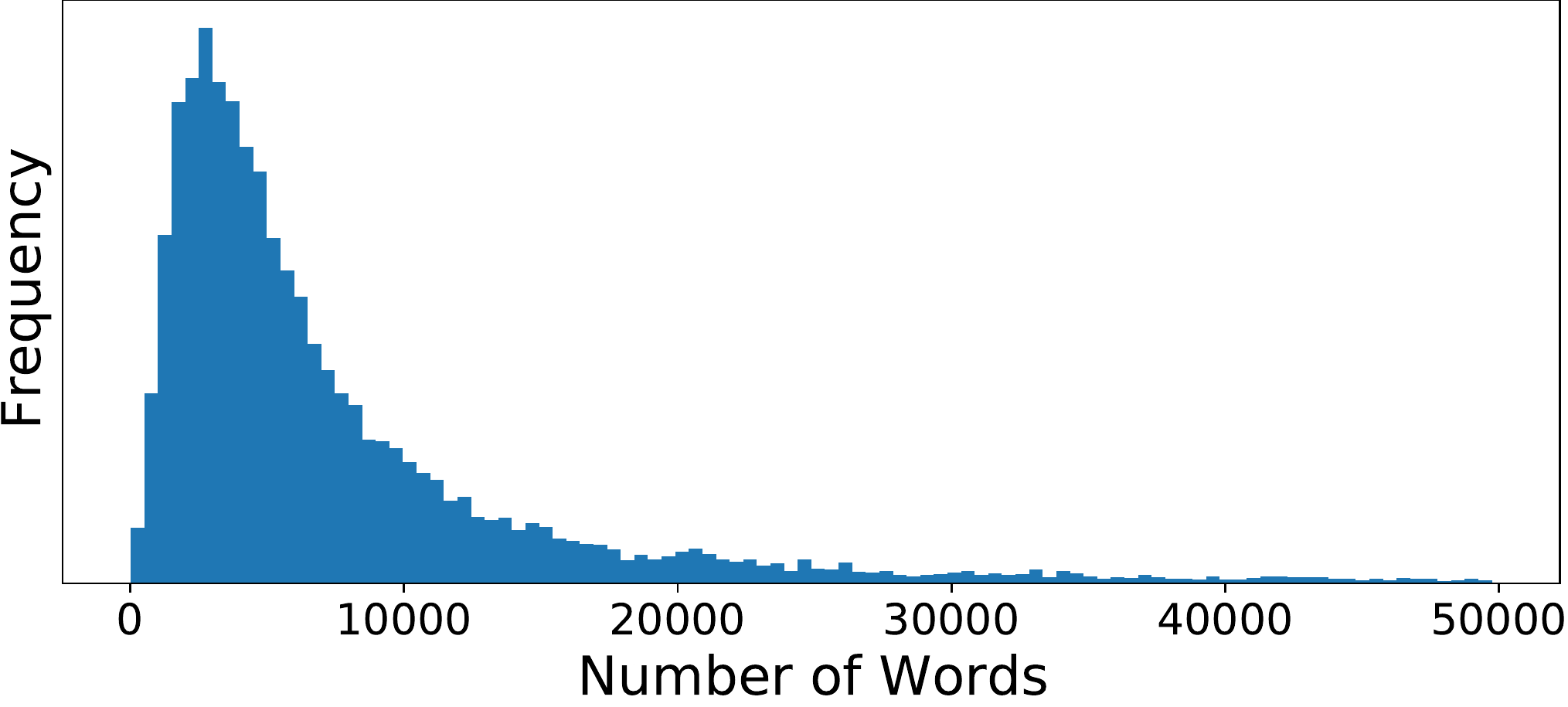}  
    \caption{Number of words per opinion.}
    \label{fig:token-per-opinion}   
\end{figure}

Court opinions are typically large documents, we can illustrate this by showing the distribution of the number of words per opinion in Figure~\ref{fig:token-per-opinion}.

\section{Highlight Extraction}
\label{section:highlight-extraction}

In this section, we will formally define the task of highlight extraction, then describe the empirical experiments we conducted and their results.

\subsection{Task Description}

We formally define the task of highlight extraction as follows.
Given a set of opinions $V$, where each opinion $v \in V$ is associated with an opinion text $T_v$ split into a set $S_v$ of $n_v$ sentences.
The task of highlight extraction is to predict a subset of sentences $\widehat{S_v} \subseteq S_v$ which will be cited in the future.
The input of this system is the text of a single opinion. It is implied that the system's parameters are learned from the complete collection of opinions, and the system's training should not be restricted to document level tasks.
Thereby, the task of identifying highlights, when formulated at sentence level can be considered as a sentence binary classification (ranking) problem.

In this work, we define this task as a sentence highlighting task. The text spans we will consider are sentences from the opinion, we want to build  model that predicts which sentence(s) in a given opinion might be cited, considering the historical observations as a ground truth that can be used for training.

This task involves the detection of network and timeline effects, as novelty arises from being the first to make a certain argument, or the first to legally qualify some facts as being compliant or not with generic statutes. As such we do expect it to be solvable at document level, at text level, if and only if we can assume that the opinion drafters are aware of the novelty of elements in the opinion, and draft them textually in a way that the task can be reduced to a Language Modeling task.

\subsection{Evaluation Metrics}
In line with the formulation of highlight extraction as a sentence ranking task, we will evaluate the different models with ranking metrics, such as Precision at 1 (P@1), Precision at R (P@R), Mean Average Precision (MAP)  and Mean Reciprocal Rank (MRR). 

In the context of supporting the work of a legal practitioner, we value the presence of relevant sentences early in the ranked list of sentences. P@1 measures the capacity of placing a highlight sentence at the top of the ranked list, P@R is a precision metric suitable for queries with a diverse number of relevant answers, MAP balances equally important metrics precision and recall and MRR is an indicator of the effort of the system's user, in relation to the rank of the first relevant sentence.

We also consider ROUGE~\cite{Lin2004}, a family of metrics for summary evaluation. For each opinion, from the list of ranked sentences, we consider the hypothesis summary to be made of the top 5 sentences. The reference summary of an opinion is made of the highlight sentences in the opinion. ROUGE will score favorably a ranker which identified sentences lexically similar to the reference sentences.

\subsection{Data}
We use a sample out of VerbCL Highlights as training material. We make use of a random sample of 88k opinions, from the pool of opinions that have at least one sentence marked as highlight. The data is split between training material and test material. The train collection contains circa 70k opinions.

As we have to respect the sequence length limitations of the deep neural models, we restricted the test collection to the opinions for which there is at least one highlighted sentence in the first $N=512$ words of the opinion, and we consider the original text as made of these N first words. This test collection contains circa 1,000 opinions.
Since the data is split based on opinions, the data in the test set is unseen for a model trained on the train set. 

As we focus on identifying highlights, we make use of documents for which we observe that they contain highlights. We leave apart in this work the task of deciding whether or not an opinion contains highlight. From the perspective of the highlight extraction task, we consider having an oracle that has already stated the existence of highlights in the dataset we use.

\begin{table*}[h]
    \begin{tabularx}{\linewidth}{|X|r|r|r|r|r|r|r|}
        \hline
        \textbf{Model} & \textbf{P@1} & \textbf{MAP} & \textbf{P@R} & \textbf{MRR} & \textbf{Rouge1-F} &  \textbf{Rouge2-F} &  \textbf{RougeL-F} \\
        \hline
        DistilBERT & 0.05 & 0.11 & 0.05 & 0.12 & 0.31 &  0.20 &  0.26 \\
        TextRank & 0.14 & 0.23 & 0.14 & 0.24 &  0.41 &  0.36 &  0.38 \\
        PreSumm & 0.31 & 0.37 & 0.30 & 0.38 &  0.52 &  0.48 &  0.48  \\
        \hline
    \end{tabularx}
    \caption{Results for the Highlight Extraction Baselines}
    \label{table:highlight-results}
\end{table*}

\subsection{Baselines}

Considering the highlight task as a sentence ranking task, we propose two different types of baseline using
\begin{itemize}[]
    \item binary classification at a sentence level;
    \item extractive summarization at a document level.
\end{itemize}

While it makes sense to consider the binary classification task, we argue that this approach will fail in this case, as most of the signals that a sentence in a legal opinion might be of interest for citation lie outside of the sentence itself. We produce this experiments to verify our assumptions that the sentence itself does not contain the signal of its importance.

An extractive summarization model will have to take the whole document into account in order to rank the sentences. On the one hand, deterministic methods will focus on one aspect for the importance of a sentence within a document. On the other hand, trainable models can be fine-tuned towards golden summaries, based on sentence-level annotations. We will use both types of models with our annotated data from VerbCL Highlights, and observe whether our annotations for importance can be learned this way. The assumption we want to confirm is that the document contains more information about the important sentence than the sentence itself, but still not enough information to lead to an accurate extraction.

We select the following models as our baselines:
\begin{itemize}[]
    \item \textbf{DistilBERT}~\cite{sanh2020distilbert}: a generalized language model that we fine-tune for the binary classification task at sentence level. We use the \texttt{transformers} library from HuggingFace~\footnote{\url{https://huggingface.co/transformers/}}. We derive a sentence ranker from this model by considering the estimated probability of belonging to the positive class as the sentence score.
    
    \item \textbf{TextRank}~\cite{barrios2016variations}: an extractive summarization deterministic algorithm. Although it can not be considered as state of the art, TextRank has proven to be a reliable baseline for summarization with a reasonable computation footprint. Considering the task of highlight detection as an extractive summarization task, we evaluate how well the standard summarization approach that ranks sentences by their importance is able to capture what is considered salient from the perspective of citing an opinion. Using the Python implementation of \texttt{PyTextRank}\cite{PyTextRank}, we retrieve the score of each sentence.
    
    \item \textbf{PreSumm}~\cite{presumm}: an extractive summarization model, based on Bert~\cite{devlin-etal-2019-bert}. PreSumm has the capacity to be fine-tuned on any dataset annotated at sentence level. PreSumm implementation~\footnote{\url{https://github.com/nlpyang/PreSumm}} has been adapted to our dataset, and we retrieve the score of each sentence after inference. The model was trained for 50k steps, which took approximatively 14 hours with 4 nVidia GPUs.
\end{itemize}

\subsection{Results}

We present our results in Table~\ref{table:highlight-results}. They confirm the initial assumptions with regard to the difficulty of the task. We formulate the highlight extraction as a task where the relevant context is the context of the current jurisprudence, therefore we make the assumption that this task can only be solved at corpus level, and not at document level. We observe from our experiences that using summarizers improves on the performance, but only to a certain degree. 

The baselines we selected are introduced in their order of complexity. The sentence ranking we derive from a binary relevance classifier based on DistilBERT performs similarly to a random ranker. This validates our assumption that highlights can not be discovered at sentence level only. 

On the other hand, the improvements shown with the summarizing models emphasizes that some of the highlights can be captured at the document level. We hypothesize that this is linked to the opinion drafter's awareness of the importance of the sentence, and its potential to be cited later. An opinion drafter is a person skilled in the legal domain, they are in position to have this intuition. They would reflect the importance of the sentence at the text level, and therefore be detected by models from the BERT lineage, who excel at isolating signals from the text~\cite{tenney2019bert,clark2019does,lin-etal-2019-open}.

The $\textrm{P}@1 = 0.31$ observed on PreSumm is a promising result, but we also have to consider that it is restricted to short documents, while a majority of documents are longer than the maximum input length. Despite the observed improvements, we conclude that none of these systems has solved the task sufficiently.  

\section{Related Work}
\label{section:related-work}

In this section, we present an overview of existing work in relation with the dataset and the tasks we want to support, from within the legal domain as well as outside of this legal domain. We provide a brief overview of the previously proposed datasets, and describe the recent advancements on the tasks of highlight extraction, citation-based summarization and case law retrieval.

It is important to note that similar tasks often arise also in patent retrieval~\cite{sarica2019engineering, hofstatter2019enriching, shalaby2019patent, shalaby2018toward, rossi2018query} and academic document retrieval domains~\cite{schaer2020overview,li2017topic,xiong2017explicit,li2017investigating}. 

When drawing parallels to the tasks outside of the legal domain, the following dimensions are of a special importance: 
\begin{itemize}[]
    \item emergence of a directed citation network;
    \item time-based dependencies (it is possible to cite only previously published sources); 
    \item highlight extraction for document summarization;
    \item predicting citations and text reuse.
\end{itemize}

We observe recent work in the field of citation analysis for scientific literature, enabled by S2ORC~\cite{lo-wang-2020-s2orc}, which we consider a similar dataset to VerbCL Citation Graph and VerbCL Highlights in content and intent. Applied to academic literature, rating novelty~\cite{hua2021extraction}, identifying contributions~\cite{hayashi2020whats} or summarizing~\cite{cachola2020tldr} share similarities with the highlight extraction task. We depart from the work done on academic literature, as we argue that the novelty is the core motivation of academic publication, while court opinions are motivated by the obligations of a public service which has to serve justice whenever it is requested. In that matter, novelty is expected to be sparser, while abundant redundancy should be the norm.

\subsection{Case Law Datasets}

CourtListener was also the original source of CaseLaw~\cite{Locke2018}, for the purpose of testing case law retrieval models and systems, while the main contribution was a test collection of 2,500 relevance assessments and an annotation tool. We observe similarly sized datasets in other languages, such as CAIL2019-SCM \cite{DBLP:journals/corr/abs-1911-08962} in Chinese. Recent meta-reviews of the field \cite{zhong-etal-2020-nlp} provide pointers to existing datasets\footnote{\url{https://github.com/thunlp/LegalPapers}}, we observe that none of them has the size required for training a complex model on a specific task. 

COLIEE~\cite{coliee2018, coliee2019} is also a relevant competition, task and dataset for legal purposes, of reduced size. It is made of multiple tasks, tackling case law retrieval with tasks 1 and 2, within the legal domain of Immigration and Citizenship in canadian courts.

Task 1 describes case law retrieval as the task of identifying which past cases should be mentioned in a query case from which citations have been removed. The rationale behind the relevance judgment, the reason why cases have been cited in the past is left unknown. Task 2 will focus on identifying relevancy at paragraph level, which is a step forward to clarifying relevance for the human reader, although the dataset does not include annotations with regard to relevance arising from either situation and facts or reasoning.

\subsection{Highlight Extraction}

Given a long document, the task of highlight extraction aims at finding the snippets of text that contain the information that a user will find useful in this document, making it a task ripe for user query biased extraction. For example, \cite{49929}, \cite{7950943} and \cite{doi:10.1080/10494820.2017.1282878} show the importance of highlighting for learning. Highlighting is described as a manual annotation task, for the benefit of meeting future information needs. The concept is put in practice also by \cite{hardy-etal-2019-highres} and \cite{cheng-lapata-2016-neural}, where highlights support the generation or the evaluation of text summaries. We observe only few attempts at creating an automated highlighting system for text, in contrast to the flourishing field of video highlight generation. 

The key concept we articulate our task around is meeting future information needs. Although we have plenty of historical data available, we can only speculate now on what is important in any opinion that is published, and what will be picked up in cases that courts will litigate in the near or far future. 

The highlight extraction task we present in this work focus on identifying the novel legal content in a new case that would potentially be cited in future cases, and evaluate how this could stem from text level analysis of the opinion, rather than identifying how an opinion would suit an argument made in another one. We refer to the Section~\ref{section:highlight-extraction} for more details.

\subsection{Case Law Retrieval}
Our work is also related to the previous work on the task of case law retrieval. In this task, a legal practitioner retrieves a list of past cases that are legally relevant to a case at hand.

COLIEE~\cite{coliee2018, coliee2019} has hosted a yearly competition including such a retrieval task, restricted to cases from the Immigration and Citizenship Courts of Canada. The target is to identify cases that could be cited in a query case. 

The task is reduced to identifying the similarities between court opinions, similarities in details of the case, in the underlying storylines and reasoning. Participants reformulated the retrieval task as a ranking task based on a pairwise relevancy classifier, with various text representation techniques applied, from lexical representations~\cite{ubirled, tran2018jnlp} to deep representations~\cite{rossi2019legal, iitpcoliee2019}. It is observed that the scarcity of data is hindering the use of advanced data-driven models. This domain relevant task has shown the difficulty to work at text level to evaluate a relevancy that professionals had annotated.

We see a potential application of our dataset in a case law retrieval task guided by arguments, where the query is an argument, and the search results a list of past opinions that could support the argument, based on the highlights that were extracted.

\subsection{Citation-based Summarization}

The task of highlight extraction finds its roots in the task of summarizing long documents. We observed many efforts in the field of summarizing by using citations, in the domain of academic literature summarization, for example. Citation-based summarization was first introduced in the context of scholarly data processing for summarizing scientific articles~\cite{IbrahimAltmami2020}.
In citation-based summarization, citation anchors (called \say{catchphrases}) for one target document over multiple citing document form a summary for this target document. The CL-SciSumm shared task~\cite{Chandrasekaran2019} aims at producing summaries of academic papers guided by citation anchors (called \say{citances}), by identifying spans of text relevant to the anchor. The approaches are evaluated against a set of golden summaries, using standard metrics of the ROUGE family~\cite{Lin2004}.

The court opinions will show a predetermined structure, as academic papers do, although the articulation in different parts or chapters will not be signaled explicitly in the text. Our task definition differs as we focus on actual verbatim quotes from cited documents which are less biased towards the writer's view of the cited document than paraphrases. Our work considers these verbatim quotes to be the target summary, as opposed to considering them as a way to build a summary similar to the golden summary.

We consider that the spans of an opinion which are later used in citations form a summary of this opinion, which a specific type of summary guided by the search for novel and important legal aspects addressed in the opinion document.
This summary characterizes what the change an opinion makes in terms of the legal argumentation, i.e., its network effect (global context), not just a summary of all the elements present in the opinion (local context).

\section{Conclusion}
\label{section:conclusion}

In this paper we introduced the task of highlight extraction from court opinions and the large dataset VerbCL of annotated court opinions aimed at supporting training and evaluation for the highlight extraction task.
Our dataset focuses on citations made in a court opinions by quoting verbatim preceding court opinions in support of a legal argument.
VerbCL is sufficiently large and can be used for training machine learning models on the highlight extraction task.
We also see the potential of VerbCL for other legal information retrieval tasks that can be informed by the citation network. 

We also demonstrated the difficulty of the highlight extraction task, which escapes the reduction to a sentence-level or document-level task.
Note that the citation network is the result of a massively distributed expert work from the date of the first citation until now.
Every citation is the result of an expert's retrieval of a relevant opinion to support the argument in the context of a specific case at hand. 
However, we can verify the importance of every specific text span from an opinion only post-hoc, i.e., not at the time of the publication but later on as we observe how the same arguments made in the opinion are subsequently re-used within other opinion documents in our dataset.
Only this reuse over time indicates the document's importance in terms of the impact it made to help shape other documents.
Future work should aim to address such network and historical effects, distilling knowledge of the previously published documents into the processing of a new one.
We believe that similar approaches will also prove useful for patent retrieval and academic search scenarios.

Future work should also consider extending the analysis to abstractive anchors, which requires a text generation task setup instead of purely retrieval-based one considered here.
Finally, considering the full document length remains a challenge for the current neural ranking approaches that should be addressed to be applicable in the legal domain. 


\begin{acks}
This research was supported by
the NWO Innovational Research Incentives Scheme Vidi (016.Vidi.189.039),
the NWO Smart Culture - Big Data / Digital Humanities (314-99-301),
the H2020-EU.3.4. - SOCIETAL CHALLENGES - Smart, Green And Integrated Transport (814961),
the Amsterdam Business School PhD program.
All content represents the opinion of the authors, which is not necessarily shared or endorsed by their respective employers and/or sponsors.
\end{acks}

\balance
\clearpage
\bibliographystyle{ACM-Reference-Format}
\bibliography{cikm_resource}


\end{document}